%% file: main.tex
\documentclass[letterpaper, 10 pt, journal]{IEEEtran}

\usepackage[T1]{fontenc}
\usepackage{times}
\renewcommand{\IEEEPARstartFONTSTYLE}{\fontencoding{T1}\fontfamily{ptm}\selectfont}
\usepackage{graphicx}
\usepackage{capt-of}
\usepackage{amsmath}
\usepackage{amssymb}
\usepackage{cite}
\usepackage{booktabs}
\usepackage{multirow}
\usepackage{array}
\usepackage{enumitem}
\usepackage[table]{xcolor}
\usepackage{tikz}
\usetikzlibrary{arrows.meta,positioning,calc}
\usepackage{algorithm}
\usepackage{algorithmic}
\usepackage{stfloats}
\usepackage[hidelinks]{hyperref}

\hypersetup{
  pdfauthor={Shuning Zhang, Shuli Zhang, Yunheng Wang, Wei Lu, Jiaxi Zhang, Yihang Kang, Tao Wang, Xiangyu Chen, Yixiao Feng, Kai Sheng, Renjing Xu},
  pdftitle={AgenticJevNav: Agentic Object Search and Navigation via Event-Driven Verification},
  pdfsubject={Event-driven navigation architecture, simulation, and physical deployment},
  pdfkeywords={agentic navigation, object search, event-driven verification, versioned evidence, legged robots}
}

\title{AgenticJevNav: Agentic Object Search and Navigation via Event-Driven Verification}

\author{Shuning Zhang, Shuli Zhang, Yunheng Wang, Wei Lu, Jiaxi Zhang,
Yihang Kang, Tao Wang, Xiangyu Chen, Yixiao Feng, Kai Sheng, and Renjing Xu}

\begin{document}
\bstctlcite{BSTcontrol}

% ---- Teaser Figure (placed after title, before abstract) ----
\IEEEaftertitletext{%
  \vspace{-0.5\baselineskip}
  \begin{center}
    \includegraphics[width=\textwidth]{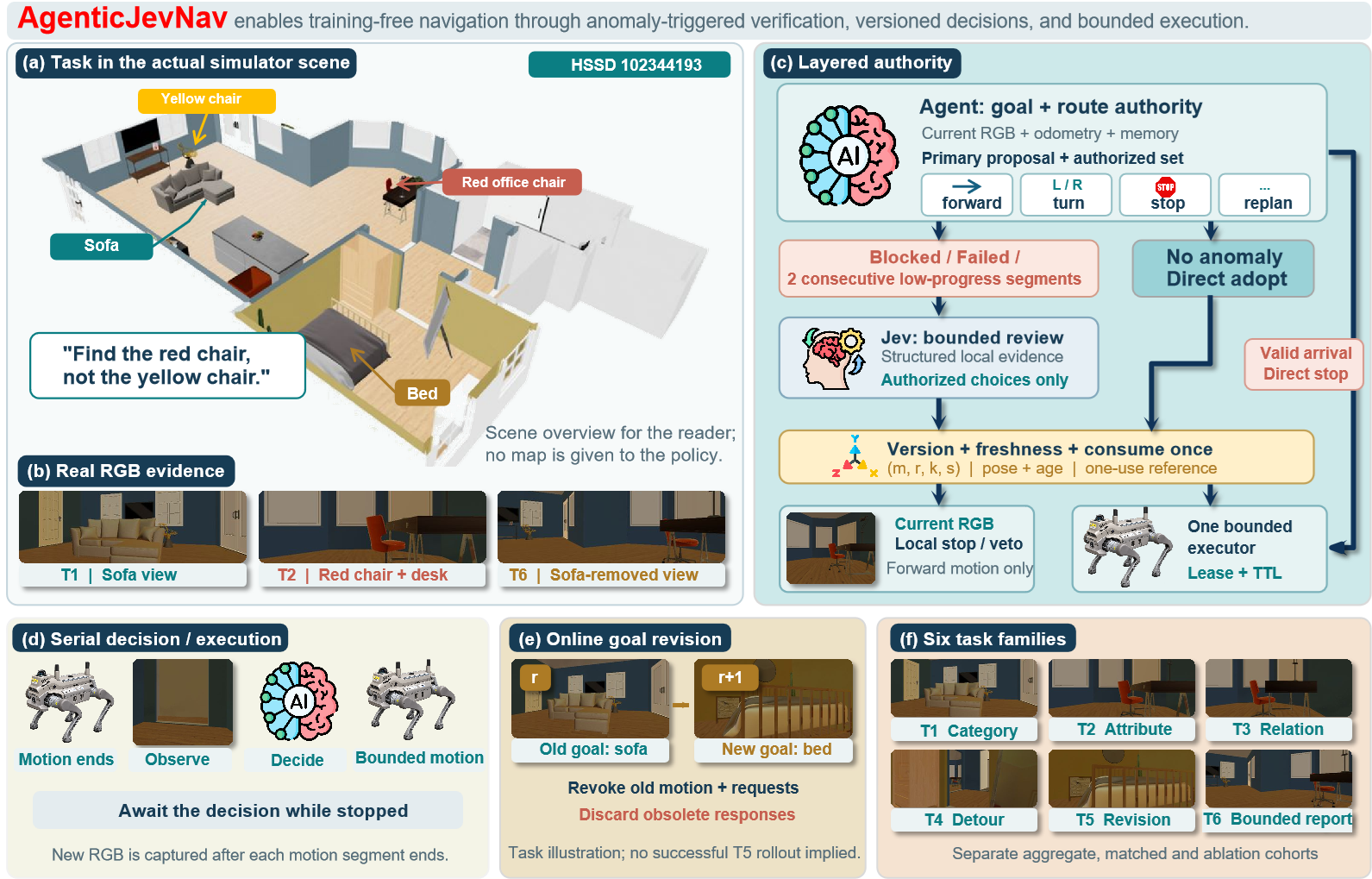}
    \vspace{-0.8\baselineskip}
    \captionof{figure}{AgenticJevNav overview. (a,b) Simulator scene, goal, and task RGB examples. The bird's-eye view is not a policy input. (c) The Agent plans and authorizes candidates. Jev reviews them after execution anomalies. Both paths pass version, freshness, and once-only checks before a single bounded executor. An independent RGB reflex can stop/veto forward motion. (d) Serial observe--decide--execute loop. (e) Goal revisions revoke obsolete requests and motion. (f) Six task families. Simulator images and illustrative robot artwork depict the setup.}
    \label{fig:teaser}
  \end{center}%
  \vspace{0.3\baselineskip}
}

\maketitle
\thispagestyle{plain}
\pagestyle{plain}

%%%%%%%%%%%%%%%%%%%%%%%%%%%%%%%%%%%%%%%%%%%%%%%%%%%%%%%%%%%%%%%%%%%%%%%%%%%%%%%%
\begin{abstract}

Agentic navigation must maintain goals despite incomplete evidence, execution failures, and delayed inference. We present AgenticJevNav, integrating goal-directed planning, evidence memory, event-triggered verification, and bounded execution. The Agent plans from RGB, odometry, memory, and feedback, retains goal ownership, and authorizes candidate actions. Jev reviews structured evidence after blocked or failed execution, or two consecutive low-progress segments. Version, freshness, and once-only checks govern execution; a single executor and independent RGB reflex constrain motion. Goal revisions revoke obsolete requests and actions. Across six simulated task families, AgenticJevNav achieves 73.33\% SR over 60 trials, 21.67 percentage points above the strongest Agent-only control. A real-world GENISOM L1 quadruped deployment achieves 70.0\% success over 30 trials, including bounded search reporting. Together, simulation and physical-robot results demonstrate the architecture's effectiveness.
\textit{Project page:} \url{http://106.55.248.254/}.
\end{abstract}


%%%%%%%%%%%%%%%%%%%%%%%%%%%%%%%%%%%%%%%%%%%%%%%%%%%%%%%%%%%%%%%%%%%%%%%%%%%%%%%%
\section{Introduction}

\IEEEPARstart{R}{esearch} on embodied navigation spans instruction following and grounding~\cite{r2r,vlmgrounder}, semantic exploration~\cite{vlfm,ovgnav}, language-model planning~\cite{navgpt,instructnav,zero_shot_planners}, and learned navigation~\cite{navid,navila}. Vision-language models ground natural-language goals in visual observations, while large language models generate structured navigation plans from task descriptions. Learned policies integrate these capabilities into end-to-end systems that map observations to actions. However, these capabilities leave a coordination problem: how should a robot maintain its goal, respond to a lack of progress, and apply delayed decisions as execution and instructions change? When perception is uncertain, when should the system trigger additional verification rather than proceeding with the Agent's initial judgment? How can the system ensure that decisions produced by remote inference remain valid and authorized when the goal has been revised or execution context has changed?

We study when verification is warranted, whether responses remain valid, and how motion can stop before remote inference returns. AgenticJevNav addresses these questions through a \emph{versioned evidence interface}. The Agent owns the goal and replans from feedback; Jev reviews only authorized actions. A shared consumer checks versions, authorization, freshness, and once-only consumption before execution. An independent RGB reflex can stop forward motion without remote inference. Here, \emph{agentic} denotes this feedback-driven planning and delegation loop within one robot. Jev has no independent goal authority.

We make three contributions to agentic navigation through coordinated planning, verification, and execution:
\begin{enumerate}
\item \textbf{Feedback-driven planning with authorized verification.}
Feedback-driven planners use environmental feedback to correct actions~\cite{innermonologue,action_contextualization}. We further specify when to trigger Jev and which authorized actions it may select. Jev reviews only after blocked or failed execution, or two consecutive low-progress segments, choosing from the Agent's authorized set. This reduces review overhead while preserving goal ownership.
\item \textbf{Versioned memory and decision-making for online goal revision.}
Interactive planning~\cite{innermonologue,instructnav} and latency mitigation~\cite{rtc} address feedback and action continuity, respectively, while semantic-map reasoning~\cite{osmag_llm} and instance-oriented navigation in dynamic environments~\cite{openin} organize semantic information for acting. We address the complementary issue of temporal validity after online goal corrections. We bind evidence to mission, revision, skill, and snapshot, and revoke stale requests upon correction. This isolates obsolete responses and maintains authorization consistency with the current task context.
\item \textbf{Execution that separates remote semantic decisions from local motion constraints.}
Inference latency affects action continuity~\cite{rtc}. We integrate RGB veto checks, skill leases, command time-to-live (TTL), and task versions within a single executor. This enables stop/veto actions without waiting for remote responses, and limits the application of invalid commands through revocation and expiry.
\end{enumerate}

%%%%%%%%%%%%%%%%%%%%%%%%%%%%%%%%%%%%%%%%%%%%%%%%%%%%%%%%%%%%%%%%%%%%%%%%%%%%%%%%
\section{Related Work}

\subsection{Foundation Models and Agentic Navigation}
VLFM~\cite{vlfm} separates semantics from geometric execution. NavGPT~\cite{navgpt} and InstructNav~\cite{instructnav} use language-model planning, whereas NaVid~\cite{navid} and NaVILA~\cite{navila} learn policies or representations. NavJev~\cite{navjev} reduces inference cost through visual compression and semantic memory. StreamVLN~\cite{streamvln} models streaming visual history. AwareVLN~\cite{awarevln} selectively triggers reasoning. Our focus is complementary: authorization, temporal validity, and once-only consumption govern whether a judgment may enter execution. Physical and visual discrepancies between simulation and real-world deployment~\cite{vlnpe} motivate our real-robot evaluation on the GENISOM L1 platform.

\subsection{Goal Grounding and Structured Decisions}
GOAT-Bench~\cite{goatbench} distinguishes category, language, and image goals. VLMaps~\cite{vlmaps} and ConceptGraphs~\cite{conceptgraphs} organize semantic evidence for navigation. LGX~\cite{lgx} uses language-model-guided exploration for zero-shot object navigation. OpenIN~\cite{openin} targets instance-oriented navigation in dynamic domestic environments, while osmAG-LLM~\cite{osmag_llm} combines semantic maps with language-model reasoning. We instead emphasize versioned evidence: rejecting obsolete judgments after goal corrections. Jev confidence measures output-distribution concentration, not calibrated perceptual truth. Valid versions do not establish target identity.

\subsection{Legged Navigation and Safety}
Perceptive locomotion~\cite{anymal} and behavior-tree navigation with online replanning and recovery~\cite{bt_recovery} support legged and wheeled-legged autonomy, while SubT systems~\cite{autonomousspot} showcase legged autonomy in extreme environments. NaVILA~\cite{navila} integrates vision-language navigation with legged platforms. Control barrier functions~\cite{amescbf} provide formal safety guarantees under model assumptions. Our local RGB reflex constrains forward motion independently of remote inference, providing neither metric clearance nor validated physical safety.

%%%%%%%%%%%%%%%%%%%%%%%%%%%%%%%%%%%%%%%%%%%%%%%%%%%%%%%%%%%%%%%%%%%%%%%%%%%%%%%%
\section{Method}

\begin{figure*}[t]
  \centering
  \includegraphics[width=\textwidth]{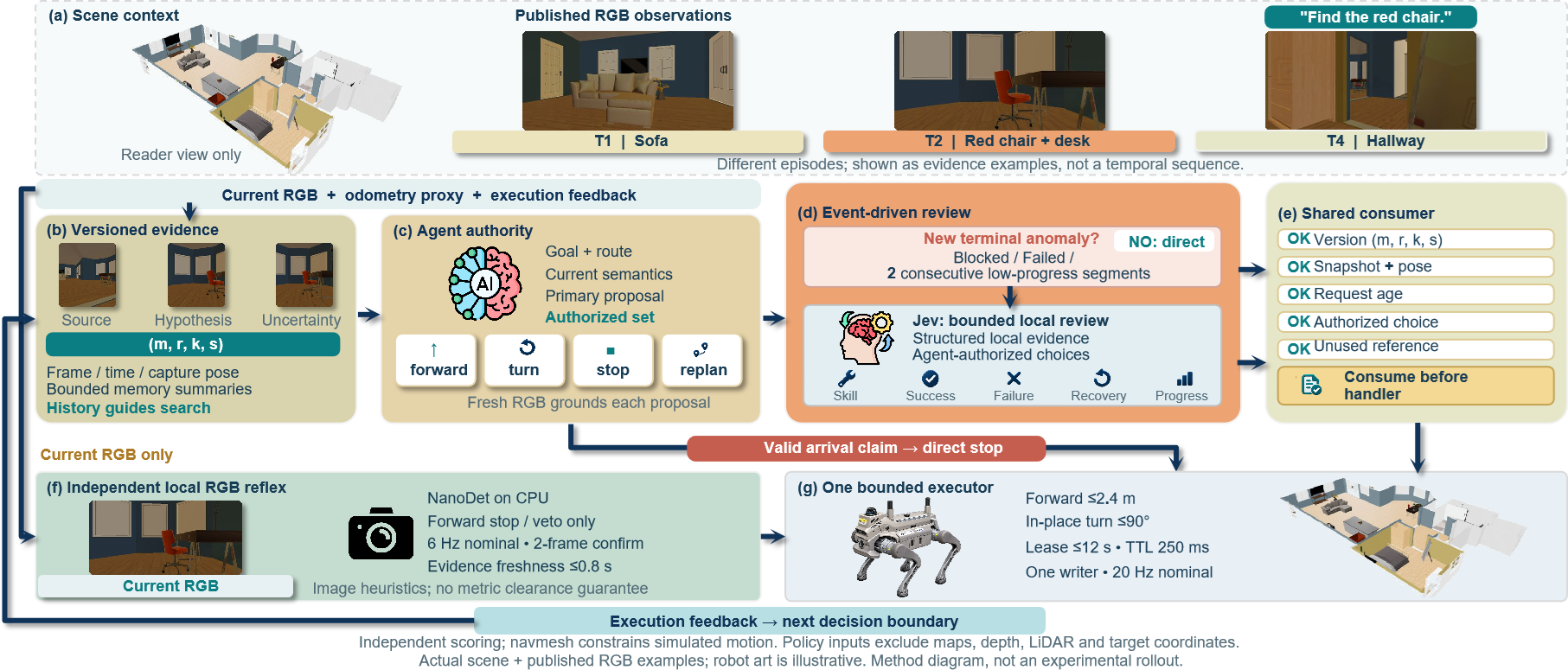}
  \caption{AgenticJevNav architecture. (a) Scene context and RGB examples from separate episodes. (b) Versioned evidence with uncertainty, timestamps, and capture poses. (c) Agent planning from RGB, odometry, memory, and feedback. (d) Direct proposal adoption or anomaly-triggered Jev review. (e) Shared version, pose, age, authorization, and once-only checks. (f) Independent CPU NanoDet-based RGB stop/veto. (g) Bounded execution with leases, command TTL, and feedback. Arrival claims use direct stop. The policy receives no maps, depth, LiDAR, or target coordinates. The navmesh supports only simulated constraints and scoring.}
  \label{fig:overview}
\end{figure*}

Figure~\ref{fig:overview} illustrates the system architecture. We formulate the task, define versioned evidence and Agent/Jev coordination, and describe local execution below.

\subsection{Problem Formulation}
\label{sec:problem}

\textbf{Task:} Given a natural-language goal $g$, budget $T$, and operational boundaries, the robot searches for the instructed instance, reaches it, and stops. In a timed search-report task, it reports \textit{not found within scope and budget} if no target is found.

\textbf{Inputs:} The policy observes $o_t=(I_t,\hat x_t,f_{t-1},z_t)$: RGB, odometry, previous-segment feedback, and local veto status. Simulation uses ideal pose-derived odometry. Hardware uses onboard odometry (Sec.~\ref{sec:hardware}). Depth, LiDAR, semantic ground truth, target coordinates, and the navigation mesh are unavailable to the policy. A goal correction increments revision $r$.

\subsection{Perception and Versioned Evidence}
\label{sec:evidence}

The shared state is
\begin{equation}
E_t=(g,r,\mathcal C_t,\hat x_t,\mathcal H_t,f_k,z_t,q_t),
\label{eq:evidence}
\end{equation}
where $\mathcal C_t$ is candidate memory, $\mathcal H_t$ is bounded decision/execution history, $f_k$ is execution feedback, $z_t$ is local veto status, and $q_t$ records provenance and freshness. Each candidate has the form
\begin{equation}
c=(\mathrm{id},\mathrm{attrs},\mathrm{rels},\mathrm{frame},\tau_{\mathrm{seen}},\mathrm{vis},\mathrm{src},\delta).
\label{eq:candidate}
\end{equation}
Here, attributes, relations, and uncertainty $\delta$ are uncalibrated RGB-model outputs. Provenance binds the source frame, timestamp, and request. Memory associates hypotheses only under small pose changes and unique box overlap. Larger viewpoint changes create unconfirmed candidates. History guides search but cannot authorize motion or arrival, which require fresh evidence. Pixel boxes are not converted to world coordinates. The Agent receives bounded memory summaries. Jev sees current semantics, proposal, odometry, and local history.

\subsection{Agent Module: Search and Route Intent}
\label{sec:agent}

The Agent outputs current-image semantics, a primary skill, and an authorized set $\mathcal A_k$, including that skill, stop, and replan. Skills are forward motion ($\leq2.4$\,m), in-place left/right turns ($\leq90^\circ$), stop, and replan. Reverse/lateral motion is disabled. Before forward motion, the target or passage-reference box center must lie in normalized horizontal interval $[0.38,0.62]$; otherwise the robot turns and observes again. Schema/numerical validation allows one same-observation correction before rejecting an invalid output.

Each action is replanned from a fresh image. An arrival claim requires a visible, centered target, a primary stop skill, an authorized set $\{\mathrm{stop},\mathrm{escalate}\}$, and zero translation and yaw commands. It revokes motion and submits the claim directly, bypassing Jev. Independent scoring determines success. Simulator settings are frozen within each cohort. Hardware arrival-policy changes are disclosed in Sec.~\ref{sec:hardware}.

\subsection{Jev Module: Bounded Judgments}
\label{sec:jev}

After an anomaly, Jev receives structured local evidence, without raw images, and returns five typed judgments: skill routing, success verification, failure detection, recovery routing, and task progress. The program validates their fields and confidence scores and combines routing, failure, and required recovery judgments. Success/progress are retained but do not certify completion. Formally,
\begin{equation}
d=\mathrm{Jev}(E_t|_{\mathrm{local}},\mathcal A_k)=\{(c_i,p_i)\}_{i\in\mathcal Q},
\label{eq:jev}
\end{equation}
where $\mathcal Q=\{\mathrm{skill},\mathrm{success},\mathrm{failure},\mathrm{recovery},\mathrm{progress}\}$, $c_i$ is a typed choice, and $p_i$ is the confidence assigned to that choice. Skill admission requires
\begin{equation}
c_{\mathrm{skill}}\in\mathcal A_k\;\land\;p_{\mathrm{skill}}\ge\tau_{\mathrm{skill}}.
\label{eq:admissible}
\end{equation}
Low confidence, replanning, or an invalid critical field rejects the proposal. If a definite failure is detected (label other than none/unknown, confidence $\geq0.85$), recovery confidence must also be at least 0.8; otherwise, the proposal is rejected. Stop/reobserve maps to stop; other recovery choices return control to the Agent. Jev cannot invent unauthorized skills or goals.

A result $d$ is bound to mission $m$, revision $r$, skill $k$, and snapshot $s$, and remains usable only if
\begin{equation}
\begin{split}
V(d)={}&[m_d=m_t]\land[r_d=r_t]\land[k_d=k_t]\\
&\land[s_d=s_t]\land\operatorname{PoseValid}(s_d,\hat x_t)\\
&\land[\Delta t_{\mathrm{request}}\leq8\,\mathrm{s}].
\end{split}
\label{eq:valid}
\end{equation}
Snapshot identity must match the consumption slot. $\operatorname{PoseValid}$ allows at most 0.8\,m translation and $35^\circ$ yaw from capture. The 8\,s age starts at coordinator-request creation, not response arrival or Agent-inference start. With version tuple $\nu=(m,r,k,s)$ and attempted-consumption references $\mathcal U$, admission further requires
\begin{equation}
\begin{split}
K(d)={}&[\mathrm{status}_d=\mathrm{accepted}]\land V(d)\\
&\land[\mathrm{id}_d\notin\mathcal U]\land[a_d\in\mathcal A_k].
\end{split}
\label{eq:consume}
\end{equation}
Accepted status means response, timestamp, and candidate-legality checks passed. $K(\bot)=0$. The consumer records the reference in $\mathcal U$ before handling and rechecks versions/execution boundaries on mutation. References cannot be reused even after handler failure. A receipt records controller-state change, not physical completion. Confidence cannot override validity or veto checks.

\subsection{Execution Module: Bounded Motion and Feedback}
\label{sec:exec}

A single executor carries mission/revision identifiers, a 12\,s skill lease, and 250\,ms command TTL. Version mismatch or budget exhaustion revokes motion. At nominal 20\,Hz it generates commands from odometric error:
\begin{equation}
\begin{aligned}
v&=\mathrm{clip}\!\bigl(k_p\,(d^{*}-\hat d),-v_{\max},v_{\max}\bigr),\\
\omega&=\mathrm{clip}\!\bigl(k_\theta\,(\Delta\theta^{*}-\widehat{\Delta\theta}),-\omega_{\max},\omega_{\max}\bigr).
\end{aligned}
\label{eq:control}
\end{equation}
Here, $d^*,\Delta\theta^*$ are skill-relative targets, and $\hat d,\widehat{\Delta\theta}$ are measured progress. We use $k_p=0.8$, $k_\theta=1.2$, $v_{\max}=0.3$\,m/s, and $\omega_{\max}=0.6$\,rad/s. Turns set $v=0$. Local skill completion requires forward progress $\geq d^*-0.06$\,m or yaw error within $4^\circ$, not task success. The executor permits only forward motion/in-place turns under navmesh constraints. Forward commands recheck RGB veto. Stop, revision, or lease expiry interrupts execution, and network waiting never renews the lease.

\subsection{Local RGB Reflex Layer}
\label{sec:reflex}

An independent CPU thread runs the OpenCV Zoo NanoDet ONNX model (OpenCV DNN, COCO classes) at nominal 6\,Hz. It checks a forward corridor $\mathcal R_{\mathrm{fwd}}$ and lower image band. A box satisfying
\begin{equation}
h(b)=\mathbf{1}\!\bigl[\,p_b\ge p_{\min}\;\land\;y_b^{(2)}\ge y_{\mathrm{thr}}\;\land\;\mathrm{overlap}(b,\mathcal R_{\mathrm{fwd}})>0\,\bigr]
\label{eq:hazard}
\end{equation}
is a hazard candidate. Same-class IoU tracking requires two-frame confirmation by default:
\begin{equation}
\mathcal B_t=\{\,b:\mathrm{count}(b)\ge n_{\mathrm{confirm}}\,\}
\label{eq:confirm}
\end{equation}
Confirmed hazards trigger forward stop/veto and feedback at the next decision boundary. The implementation also checks lateral overlap/lower-band geometry and briefly retains hazards. Missing or stale detection results veto forward motion. Turning does not require forward clearance. These image heuristics neither estimate metric distance nor guarantee whole-body collision avoidance. Nominal 6\,Hz is not stopping latency.

\subsection{AgenticJevNav Algorithm Flow}
\label{sec:coord}

Each iteration captures an observation, validates the Agent proposal, and applies the anomaly gate. Jev review is triggered by blocked or failed execution, or by two consecutive low-progress segments. Other valid proposals proceed through the shared admission checks. Goal corrections update the current version and revoke obsolete requests and motion. Algorithm~\ref{alg:loop} details the full loop.

\begin{algorithm}[t]
\caption{Agentic Planning, Review, and Execution}
\label{alg:loop}
\begin{algorithmic}[1]
\REQUIRE $g$, operational boundaries $\mathcal B$, budget $T$
\STATE $(m,r)\gets(m_0,0)$, $\mathcal C,\mathcal H,\mathcal U\gets\varnothing$, $\eta\gets\eta_0$
\WHILE{$t<T\ \land\ \neg\mathrm{Terminal}$}
\STATE $\mathrm{AwaitMotionEnd}()$, $o_t\gets\mathrm{Capture}()$
\STATE $\nu_{\mathrm{req}}\gets\nu_{\mathrm{current}}\gets(m,r,k_{\mathrm{new}},s_t)$, \\$p_t\gets\mathrm{Validate}(\mathrm{Agent}(g,o_t,\mathcal C,\mathcal H))$
\IF{$p_t=\bot\ \lor\ \nu_{\mathrm{req}}\ne\nu_{\mathrm{current}}\ \lor\ \neg\mathrm{PoseValid}(s_t,\hat x_t)$}
\STATE \textbf{continue}
\ENDIF
\STATE $(E_t,\mathcal C)\gets\mathrm{UpdateEvidence}(p_t,o_t,\mathcal C)$
\IF{$\mathrm{ArrivalValid}(p_t)\ \land\ \neg\mathrm{SearchReportTask}$}
\STATE $\mathrm{SubmitClaim}(p_t)$, $\mathrm{Terminal}\gets\mathrm{true}$
\ELSE
\STATE $(b_t,\eta)\gets\mathrm{Gate}(f_{t-1},\eta;m,r)$
\STATE $F_t\gets b_t{=}1\,?\,\lambda().\mathrm{Admit}(\mathrm{Jev}(E_t|_{\mathrm{local}},\mathcal A_k))\,:$\\$\,\lambda().\mathrm{Direct}(p_t.a,\mathcal A_k)$
\STATE $d\gets\mathrm{AwaitCoord}(\nu_{\mathrm{req}},F_t)$
\IF{$K(d)$}
\STATE $\mathcal U\gets\mathcal U\cup\{\mathrm{id}_d\}$, $\rho\gets\mathrm{Consume}(d;\nu_{\mathrm{current}},\mathcal B,T)$
\STATE $\mathcal H\gets\mathrm{Append}(\mathcal H,d,\rho)$
\ENDIF
\ENDIF
\ENDWHILE
\STATE $\mathrm{CancelMotion}()$, $\mathrm{Persist}(\mathcal H,\mathrm{trajectory},\mathrm{claims})$
\RETURN Raw trajectories, claims, and decision/execution receipts
\end{algorithmic}
\end{algorithm}

%%%%%%%%%%%%%%%%%%%%%%%%%%%%%%%%%%%%%%%%%%%%%%%%%%%%%%%%%%%%%%%%%%%%%%%%%%%%%%%%
\begin{table*}[!t]
\caption{Six-task simulator comparison: 60 trials per method, ten per task. Metrics and scoring subsets are defined in the text.}
\label{tab:main}
\begin{center}
\footnotesize
\setlength{\tabcolsep}{2pt}
\renewcommand{\arraystretch}{1.08}
\begin{tabular*}{\textwidth}{@{\extracolsep{\fill}}lccccccccc@{}}
\toprule
\rowcolor[HTML]{D9E2F3}\textbf{Method}
 & \textbf{Venue}
 & \textbf{\shortstack{SR\\(\%) $\uparrow$}}
 & \textbf{\shortstack{NE\\(m) $\downarrow$}}
 & \textbf{\shortstack{Path\\(m) $\downarrow$}}
 & \textbf{\shortstack{Time\\(s) $\downarrow$}}
 & \textbf{\shortstack{Wait\\(\%) $\downarrow$}}
 & \textbf{\shortstack{Calls\\$\downarrow$}}
 & \textbf{\shortstack{Navmesh\\constraint (\%) $\downarrow$}}
 & \textbf{\shortstack{SPL\\$\uparrow$}} \\
\midrule
\rowcolor[HTML]{F2F2F2}\multicolumn{10}{c}{\textit{Agent controls}} \\
\midrule
A0: Agent-only & --- & 51.67 & $\mathbf{0.61}_{12.9\%}$ & 2.49 & 120.40 & 69.19 & 9.20 & 30.00 & 0.469 \\
A1: Agent + rule events & --- & 40.00 & 0.84 & 2.24 & 106.85 & 67.95 & 8.78 & 18.33 & 0.474 \\
\rowcolor[HTML]{FFF2CC}\textbf{A2: AgenticJevNav (ours)} & Ours & $\mathbf{73.33}_{41.9\%}$ & 0.72 & 3.18 & 140.55 & 79.40 & 12.42 & $\mathbf{15.00}_{0.0\%}$ & $\mathbf{0.591}_{24.7\%}$ \\
\midrule
\rowcolor[HTML]{F2F2F2}\multicolumn{10}{c}{\textit{Adapted navigation methods}} \\
\midrule
VLFM~\cite{vlfm} & ICRA'24 & 0.00 & 2.35 & 5.16 & N/A & 78.74 & 71.52 & 43.33 & 0.000 \\
NaVid~\cite{navid} & RSS'24 & 23.33 & 0.85 & 2.99 & 62.88 & 32.86 & 60.85 & 20.00 & 0.289 \\
ApexNav~\cite{apexnav} & RA-L'25 & 40.00 & 2.11 & $\mathbf{1.81}_{19.2\%}$ & 91.19 & 31.02 & 54.08 & 50.00 & 0.160 \\
OmniNav~\cite{omninav} & ICLR'26 & 26.67 & 2.81 & 5.53 & 84.96 & 52.46 & 47.30 & 100.00 & 0.114 \\
JanusVLN~\cite{janusvln} & ICLR'26 & 33.33 & 0.82 & 4.77 & 77.93 & 78.23 & 64.50 & $\mathbf{15.00}_{0.0\%}$ & 0.403 \\
StreamVLN~\cite{streamvln} & ICRA'26 & 21.67 & 0.79 & 4.38 & $\mathbf{23.74}_{58.1\%}$ & $\mathbf{24.31}_{21.6\%}$ & 67.78 & 30.00 & 0.232 \\
AwareVLN~\cite{awarevln} & CVPR'26 & 13.33 & 0.97 & 8.05 & 56.68 & 43.02 & 74.13 & 38.33 & 0.084 \\
NavJev~\cite{navjev} & arXiv'26 & 31.67 & 0.70 & 5.42 & 72.34 & 73.33 & $\mathbf{5.65}_{35.6\%}$ & 36.67 & 0.269 \\
\bottomrule
\end{tabular*}
\par\vspace{5pt}
\begin{minipage}{\textwidth}
\footnotesize\raggedright
\textbf{Metrics:} Bold: column best; subscripts: relative improvement over the runner-up (0.0\% for ties). SR weights tasks equally. Success requires a final pose within 1.6\,m of the target axis-aligned bounding box and heading error $\leq60^\circ$; T6 requires true absence and a valid not-found report within budget. Time averages successful trials only. NE is final geodesic distance to a valid stopping region over T1--T5. SPL covers static T1--T4~\cite{anderson2018evaluation}. Path accumulates translation through termination.

\end{minipage}
\end{center}
\end{table*}

\section{Simulation Experiments}

We evaluate three research questions\footnote{Interactive tables, per-trial GIFs, and 3D replays: \url{http://106.55.248.254/all-results.html}.}:
\begin{enumerate}
    \item \textbf{RQ1:} Can event-triggered Jev review of Agent proposals substantially improve navigation success by enabling recovery from execution anomalies?
    \item \textbf{RQ2:} How do review timing, consumption authority, evidence richness, and RGB reflex enforcement affect navigation outcomes?
    \item \textbf{RQ3:} Which grounding, search, and arrival failures remain, and what do execution traces reveal?
\end{enumerate}

\subsection{Experimental Setup}

Six tasks test category search, attributes, relations, detouring, goal revision, and bounded absence reporting. Each task runs with a 300\,s budget.

\begin{itemize}[leftmargin=*,noitemsep,topsep=2pt]
  \item \textbf{T1} (exploration): find the sofa outside the initial view, testing category search and discovery.
  \item \textbf{T2} (attribute grounding): find the red chair, not the yellow chair. Two seats are present; attributes are verified in the rendered scene.
  \item \textbf{T3} (relational evidence): find the chair beside the table, not the sofa. Two seats are present; the table relation distinguishes the target.
  \item \textbf{T4} (detouring): find the sofa via a detour route. Paired blocked/clear layouts require adaptive path recovery.
  \item \textbf{T5} (goal revision): switch from sofa to bed 1\,s after motion begins, testing online goal revision and replanning.
  \item \textbf{T6} (absence reporting): sofa possibly absent. Paired present/removed scenes; success requires a valid not-found report within budget.
\end{itemize}

\textbf{Runtime:} AgenticJevNav uses pretrained components without navigation-specific weight updates. The Agent requests GPT-6-astra with low reasoning effort. Jev is version 1.13.0, with routing/success/failure/recovery/progress thresholds 0.35/0.80/0.85/0.80/0.75. The coordinator timeout is 19\,s. Result/consumption age limits are 8\,s. The reflex uses NanoDet-Plus-m-1.5x, $416\times416$ input, nominal 6\,Hz CPU inference, two-frame confirmation, and 0.8\,s freshness. Motion settings follow Sec.~\ref{sec:exec}.

\subsection{Baselines}

A0 directly adopts Agent proposals. A1 applies frozen local rules. A2 invokes Jev on anomalies and otherwise adopts the proposal. The Agent controls share RGB, odometry, memory, serial scheduling, executor, local checks, and direct arrival stop. We also adapt VLFM~\cite{vlfm}, ApexNav~\cite{apexnav}, OmniNav~\cite{omninav}, AwareVLN~\cite{awarevln}, NavJev~\cite{navjev}, NaVid~\cite{navid}, StreamVLN~\cite{streamvln}, and JanusVLN~\cite{janusvln}.

\subsection{Navigation Success with Jev Review (RQ1)}
\label{sec:main_results}

Table~\ref{tab:main} compares AgenticJevNav (A2) with two Agent controls and eight adapted navigation methods over 60 trials each (ten per task). \textbf{The results answer RQ1 positively: event-triggered Jev review is associated with a substantial improvement in navigation success.} A2 completes 44/60 trials (73.33\% SR), compared with 31/60 for direct proposal adoption (A0, 51.67\%) and 24/60 for frozen rule-based handling (A1, 40.00\%). This represents 13 and 20 additional successful trials, respectively, with gains of 21.67 and 33.33 percentage points. Relative to the strongest Agent control, A2 improves SR by 41.9\%, showing the practical value of augmenting goal-directed planning with bounded, context-dependent review.

\textbf{Navigation effectiveness.} A2 ranks first in both SR and SPL among all 11 methods. Its 73.33\% SR exceeds the strongest adapted baseline, ApexNav (40.00\%), by 33.33 percentage points. Its SPL of 0.591 improves on the runner-up A1 (0.474) by 24.7\%, indicating that the success advantage also translates into better success-weighted path efficiency. A2 attains 0.72\,m NE and ties JanusVLN for the lowest navmesh-constraint rate (15.00\%), compared with 30.00\% for A0 and 18.33\% for A1. Although A2 does not minimize NE or raw travel distance, it more often converts navigation attempts into completed tasks; short travel alone can instead reflect early termination without success.

\textbf{Selective review and cost.} The architectural distinction is actionable review: after an execution anomaly, Jev can redirect an authorized action rather than simply accept the Agent's proposal or apply a frozen rule. The versioned interface checks that the resulting decision remains current and authorized before execution. Jev accounts for only 0.19\% of runtime, whereas Agent waiting accounts for 79.40\%, consistent with selective rather than continuous review. The overall system nevertheless incurs higher mean time on successful trials (140.55\,s versus 120.40\,s/106.85\,s for A0/A1) and more calls per trial (12.42 versus 9.20/8.78). Thus, AgenticJevNav's main advantage is higher completion success and SPL with explicitly measured overhead, not minimum latency or travel. These results support the effectiveness of its agentic coordination architecture; RQ2 further examines the contributions of review timing, authority, and evidence.

\subsection{Ablations by Design Axis (RQ2)}

Table~\ref{tab:ablation} reports five 60-trial ablations across four design axes. Full A2 achieves the highest SR (73.33\%) and SPL (0.591) among all configurations. SPL follows Table~\ref{tab:main}, averaging the 40 static T1--T4 trials; SR, Wait, and Calls cover all 60 trials.

\textbf{Trigger timing.} Event-triggered Jev achieves 73.33\% SR and 0.591 SPL with 12.42 calls per trial. A 5\,s periodic timer produces only two action changes, raises calls to 18.53, and reduces SR/SPL to 65.00\%/0.557. Targeted review yields better navigation outcomes with fewer calls than periodic review.

\textbf{Consumption authority.} Advisory-only Jev expends 13.25 calls per trial without influencing any action, reaching 71.67\% SR and 0.525 SPL. Full A2 improves both metrics with fewer calls, supporting the value of allowing authorized review decisions to block or redirect execution rather than merely generating assessments.

\textbf{Evidence richness.} Full A2 supplies Jev with instance memory and decision history. Removing instance memory yields the lowest SR/SPL (61.67\%/0.394), whereas using only the current view reaches 66.67\%/0.488. Both remain below full A2, supporting the use of structured historical evidence for effective review.

\textbf{Reflex enforcement.} Disabling the local RGB reflex reduces SR/SPL to 65.00\%/0.506, despite fewer calls (10.62). The shadow monitor records 1,709 would-veto events and 16 navmesh-constraint trials, with no compensating Jev requests. These observations support combining immediate RGB checks with deliberative review: lower inference cost alone does not preserve navigation effectiveness.

\begin{table}[!t]
\caption{Full A2 and five ablations (60 trials each).}
\label{tab:ablation}
\begin{center}
\footnotesize
\setlength{\tabcolsep}{1.5pt}
\begin{tabular}{lcccc}
\toprule
\rowcolor[HTML]{D9E2F3}\textbf{Configuration} & \textbf{SR (\%)} $\uparrow$ & \textbf{Wait (\%)} $\downarrow$ & \textbf{Calls} $\downarrow$ & \textbf{SPL} $\uparrow$ \\
\midrule
\rowcolor[HTML]{FFF2CC}Full A2 (Table~\ref{tab:main}, reused) & \textbf{73.33} & 79.40 & 12.42 & \textbf{0.591} \\
\midrule
\multicolumn{5}{l}{\textit{Jev trigger timing}} \\
A2 timer-gated Jev (5\,s) & 65.00 & 81.83 & 18.53 & 0.557 \\
\midrule
\multicolumn{5}{l}{\textit{Jev consumption}} \\
A2 advisory-only (not consumed) & 71.67 & \textbf{79.07} & 13.25 & 0.525 \\
\midrule
\multicolumn{5}{l}{\textit{Evidence richness}} \\
A2 without instance memory & 61.67 & 80.50 & 15.52 & 0.394 \\
A2 current view only & 66.67 & 81.07 & 13.15 & 0.488 \\
\midrule
\multicolumn{5}{l}{\textit{Reflex enforcement (simulation)}} \\
A2 without local RGB reflex & 65.00 & 79.35 & \textbf{10.62} & 0.506 \\
\bottomrule
\end{tabular}
\par\vspace{3pt}
\end{center}
\end{table}

% Keep float and paragraph gaps natural on the remaining figure-heavy pages.
\raggedbottom
\subsection{Failure-Source Analysis (RQ3)}
\label{sec:qualitative}

\begin{figure}[t]
  \centering
  \input{figures/editable/fig06_failure_breakdown.tex}
  \caption{A2 failure analysis for the same 60-trial cohort as Table~\ref{tab:main} (44 successes, 16 failures). (a) Mutually exclusive recorded failure outcomes, their task composition, and their percentages among all failures. (b) Final distance to the target bounding box and heading error; shading marks the region satisfying both SR thresholds ($\leq1.6$\,m, $\leq60^\circ$). Three T3 completion claims cluster at 1.828--1.830\,m; one T2 trial fails heading only.}
  \label{fig:failure}
\end{figure}
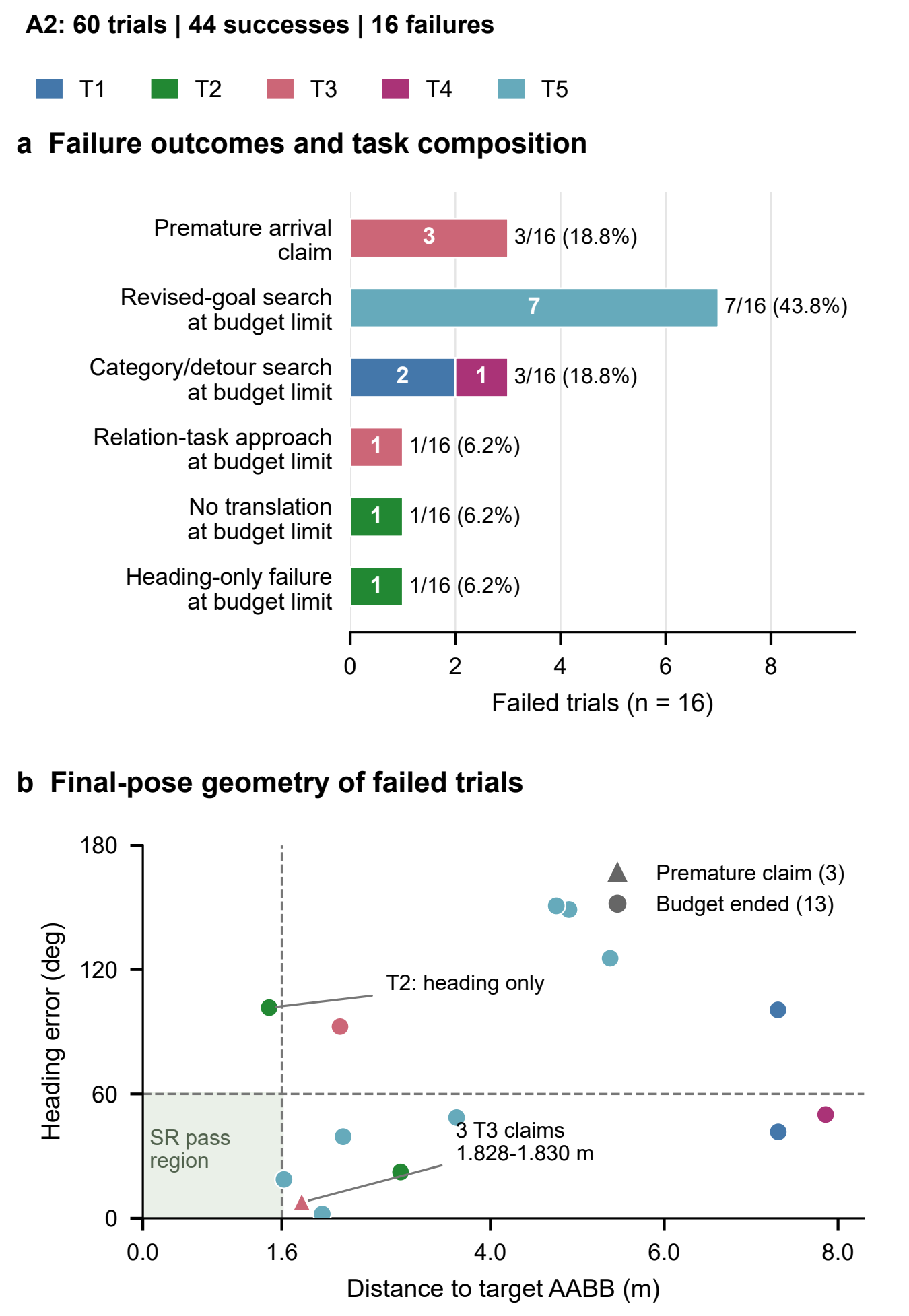

Figure~\ref{fig:failure} analyzes all 16 failures from the A2 cohort in Table~\ref{tab:main}: T1/T2 contribute two each, T3 four, T4 one, and T5 seven; T6 has none. Categories summarize recorded outcomes rather than independently established causes.

\textbf{Premature arrival claims (3 failures).} Three T3 trials declare completion at target distances of 1.828--1.830\,m and heading errors of 7.30--7.92$^\circ$. Their headings satisfy the criterion, but their distances exceed 1.6\,m, exposing insufficient approach distance. These geometric checks alone do not establish correct instance grounding.

\textbf{Budget-ended failures (13 failures).} Seven occur during revised-goal search (T5), three during category/detour search (two T1, one T4), and one during a T3 approach. The remaining two T2 trials record no translation and a heading-only failure, respectively. Recorded Agent waiting spans 134.83--262.44\,s; six trials record RGB reflex trips and three include Jev requests. These events document interventions during unsuccessful trials. One T5 trial ends only 0.024\,m beyond the distance threshold, while the heading-only T2 trial ends at 1.453\,m with 101.66$^\circ$ error. The outcomes motivate better approach/stopping decisions and revised-goal search under inference latency.

\input{physical_results}

%%%%%%%%%%%%%%%%%%%%%%%%%%%%%%%%%%%%%%%%%%%%%%%%%%%%%%%%%%%%%%%%%%%%%%%%%%%%%%%%
\section{Discussion and Limitations}

\textbf{Agency and grounding.} Explicit goal ownership, review, and action authority do not establish target identity or arrival. Correct goal revision alone does not ensure search progress; grounding and approach planning remain limitations.

\textbf{Verification and latency.} More frequent review increases cost without reliable completion gains. Execution checks may still reject a routed skill. Optimization must balance calls, consumed interventions, and progress; primary-model waiting dominates the serial loop.

%%%%%%%%%%%%%%%%%%%%%%%%%%%%%%%%%%%%%%%%%%%%%%%%%%%%%%%%%%%%%%%%%%%%%%%%%%%%%%%%
\section{Conclusion}

AgenticJevNav separates goal ownership, delegated review, and bounded execution. The Agent maintains the goal and plans from memory and feedback; Jev reviews authorized candidates after execution anomalies. A versioned interface validates decisions and bounds actions, allowing goal revisions to revoke obsolete requests and motion.

The system achieves 73.33\% SR across 60 simulated trials, exceeding the strongest Agent-only control by 21.67 percentage points, and 70.0\% (21/30) in one physical setting. Agent controls and ablations characterize review's cost--progress tradeoff. The results support auditable agentic coordination with explicit authority, decision validity, and execution limits.

%%%%%%%%%%%%%%%%%%%%%%%%%%%%%%%%%%%%%%%%%%%%%%%%%%%%%%%%%%%%%%%%%%%%%%%%%%%%%%%%

\bibliographystyle{IEEEtran}
\bibliography{references}

\end{document}

%% file: figures/editable/fig06_failure_breakdown.tex
% Current Table I cohort; rebuild with build_current_failure.py.
\includegraphics[width=\linewidth]{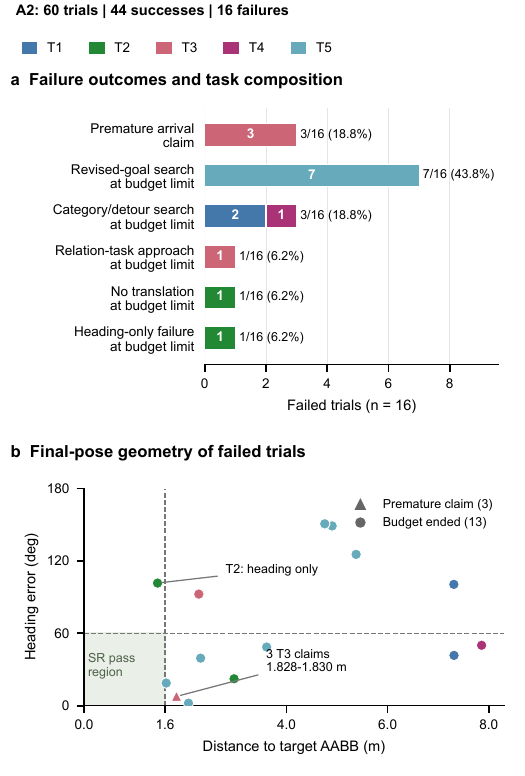}

%% file: physical_results.tex
\section{Real-Robot Experiments}
\label{sec:hardware}

\begin{figure*}[t]
  \centering
  % ---- top row: Figure 4 (scene photo) and Table 3 side by side ----
  \begin{minipage}[t]{\columnwidth}
    \vspace{0pt}%
    \centering
    \includegraphics[width=\linewidth]{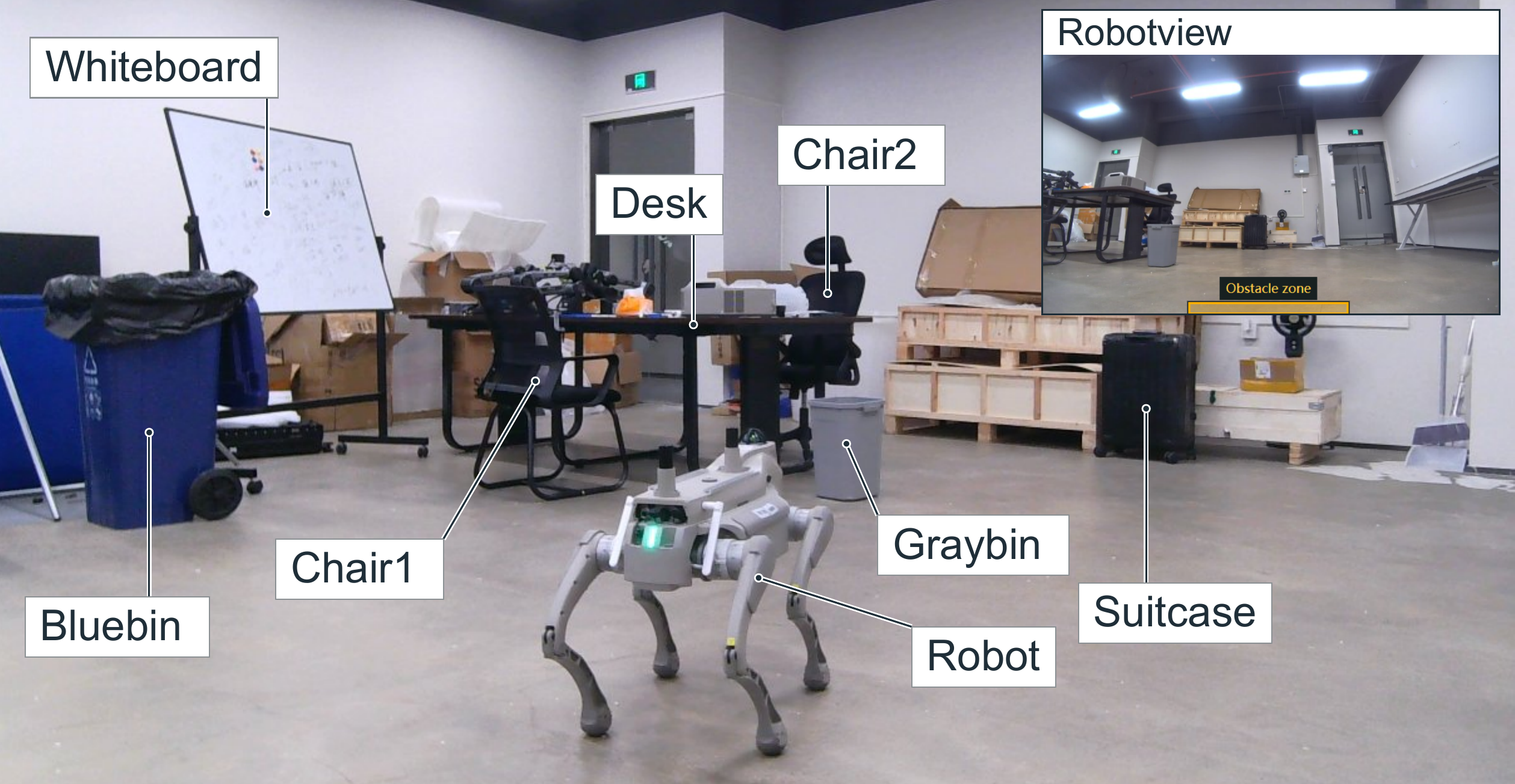}
    \captionof{figure}{Indoor quadruped setting with task objects annotated. Inset: robot RGB with the supplied obstacle-zone overlay. The scene view is not a policy input.}
    \label{fig:hardware}
  \end{minipage}
  \hfill
  \begin{minipage}[t]{\columnwidth}
    \vspace{0pt}%
    \centering
    \captionof{table}{Physical A2 results: five trials per task in one setting; R6 scores bounded reporting.}
    \label{tab:hardware}
    \footnotesize
    \setlength{\tabcolsep}{1.5pt}
    \begin{tabular*}{\linewidth}{@{\extracolsep{\fill}}lrrrr@{}}
    \toprule
    \rowcolor[HTML]{D9E2F3}\textbf{Task} & \textbf{SR (\%)} & \textbf{Time (s)} & \textbf{Agent calls} & \textbf{Jev calls} \\
    \midrule
    R1 Category & 60.0 & 142.3 & 6.8 & 0.4 \\
    R2 Attribute & 40.0 & 198.4 & 10.4 & 0.6 \\
    R3 Relation & 80.0 & 195.5 & 8.8 & 0.0 \\
    R4 Detour & 80.0 & 224.7 & 9.8 & 0.6 \\
    R5 Revision & 60.0 & 192.8 & 9.6 & 0.2 \\
    R6 Report & 100.0 & 223.9 & 9.2 & 0.0 \\
    \midrule
    Overall & 70.0 & 196.3 & 9.1 & 0.3 \\
    \bottomrule
    \end{tabular*}
    \par\vspace{3pt}
    {\scriptsize Task means cover five trials; overall means cover all 30. Time is session duration, including cleanup, cancellation, and post-report waiting. Agent and Jev calls count requests, including retries (273 Agent and 9 Jev calls in total).}
  \end{minipage}

  \medskip

  % ---- bottom row: Figure 5 (full width) ----
  \includegraphics[width=\textwidth]{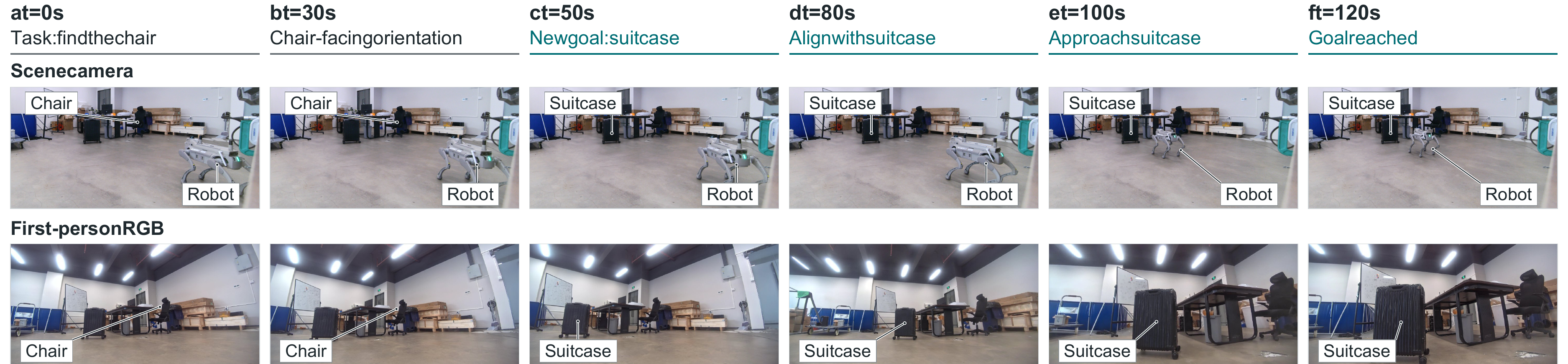}
  \captionof{figure}{Successful R5 goal revision (trial 5). Top/bottom rows show scene/robot views at approximately 0, 30, 50, 80, 100, and 120\,s. Labels identify the robot and chair/suitcase. The goal changes from chair to suitcase at 22.7\,s; panel (c) shows the revised goal, not its issuance. The robot then aligns, approaches, and reaches the suitcase. Timestamped JPEGs use fixed camera crops; times denote frame reception.}
  \label{fig:physical_sequence}
\end{figure*}

\subsection{Platform and Deployment}
Full A2 runs on a GENISOM L1 quadruped\footnote{Physical-robot trials and media: \url{http://106.55.248.254/physical/}.} with Jetson Orin NX 16\,GB navigation compute and RK3588 motion control. The policy uses front RGB and onboard \texttt{/odom/current\_pose}, retaining the same Agent/Jev interface and single executor. RGB is 1920$\times$1080, decoded to 960$\times$540; a separate scene camera records context only.

Forward requests specify 0.1--2.0\,m of odometric travel. Limits are 0.5\,s command TTL (simulation: 250\,ms), 500\,ms pose age, 90\,s per segment capped by remaining budget, and 10\,s without progress. The 0.05 command cap uses normalized SDK units, not m/s. RGB checks run at nominal 6\,Hz; odometry, stopping distance, and latency remain uncalibrated.

\subsection{Tasks and Trial Protocol}
We conducted five trials per task in the setting in Fig.~\ref{fig:hardware}: R1 chair search; R2 larger-bin selection; R3 table-side rather than whiteboard-side chair selection; R4 chair approach around a small bin; R5 chair-to-suitcase revision after motion begins; R6 red-chair search with non-detection reporting.

R1--R5 require reaching the instructed target without a task-invalidating collision; a timeout fails unless the record confirms completion. R6 requires reporting non-detection within the searched scope after the search cutoff, rather than proving absence.

Figure~\ref{fig:physical_sequence} shows successful chair-to-suitcase revision and approach in R5.5. The robot first faces the chair (panels a--b). After the goal changes to the suitcase at 22.7\,s, it aligns with the suitcase (panel d), approaches it (panel e), and stops with the suitcase filling the first-person view (panel f).

\subsection{Results and Failure Analysis}
A2 succeeds in 21 of 30 trials (70.0\%; Table~\ref{tab:hardware}). R2 has the lowest success rate (2/5), with errors in comparative grounding. All five R6 reports satisfy the bounded reporting rule. Mean session time is 196.3\,s, with 9.1 Agent requests and 0.3 Jev requests per trial.

The nine failures comprise two R1 collisions, two R2 wrong arrivals at a water bucket, four timeouts (R2.2, R3.5, R4.3, R5.4), and one visual-monitor failure (R5.2). No single cause accounts for them. Together, they point to limits in grounding, approach, and runtime.

Two events illustrate review and version control. In R4.1, RGB veto feedback triggered Jev's selection of an authorized right turn. In R5.5, version checks rejected an in-flight Agent response after a version change during termination. R5 trials retained the corrected goal without resetting the budget. These traces document coordination without isolating its effect on success. R6 reports occurred 200.22--210.72\,s after session start; session time includes post-report waiting.